\pdfoutput=1

\documentclass[11pt]{article}

\usepackage[]{ACL2023}

\usepackage{times}
\usepackage{latexsym}

\usepackage[T1]{fontenc}

\usepackage[utf8]{inputenc}

\usepackage{microtype}

\usepackage{inconsolata}
\usepackage{graphicx}
\usepackage{caption}
\usepackage{subcaption}
\usepackage{paralist}
\usepackage{hyperref}       
\usepackage{url,multirow}            
\usepackage{booktabs}       
\usepackage{amsfonts}       
\usepackage{nicefrac}       

\title{Zero-Shot Prompting for Implicit Intent Prediction and Recommendation\\ with Commonsense Reasoning}

\author{Hui-Chi Kuo\quad Yun-Nung Chen \\
  National Taiwan University, Taipei, Taiwan \\
  \texttt{r09922a21@csie.ntu.edu.tw\quad y.v.chen@ieee.org} \\}

\begin{document}
\maketitle
\begin{abstract}
The current generation of intelligent assistants require explicit user requests to perform tasks or services, often leading to lengthy and complex conversations. In contrast, human assistants can infer multiple implicit intents from utterances via their commonsense knowledge, thereby simplifying interactions.
To bridge this gap, this paper proposes a framework for multi-domain dialogue systems. This framework automatically infers implicit intents from user utterances, and prompts a large pre-trained language model to suggest suitable task-oriented bots.
By leveraging commonsense knowledge, our framework recommends associated bots in a zero-shot manner, enhancing interaction efficiency and effectiveness. 
This approach substantially reduces interaction complexity, seamlessly integrates various domains and tasks, and represents a significant step towards creating more human-like intelligent assistants that can reason about implicit intents, offering a superior user experience.\footnote{Code: \url{http://github.com/MiuLab/ImplicitBot}.}

\end{abstract}

\section{Introduction}

\begin{figure*}[t!]
  \centering
  \includegraphics[width=\linewidth]{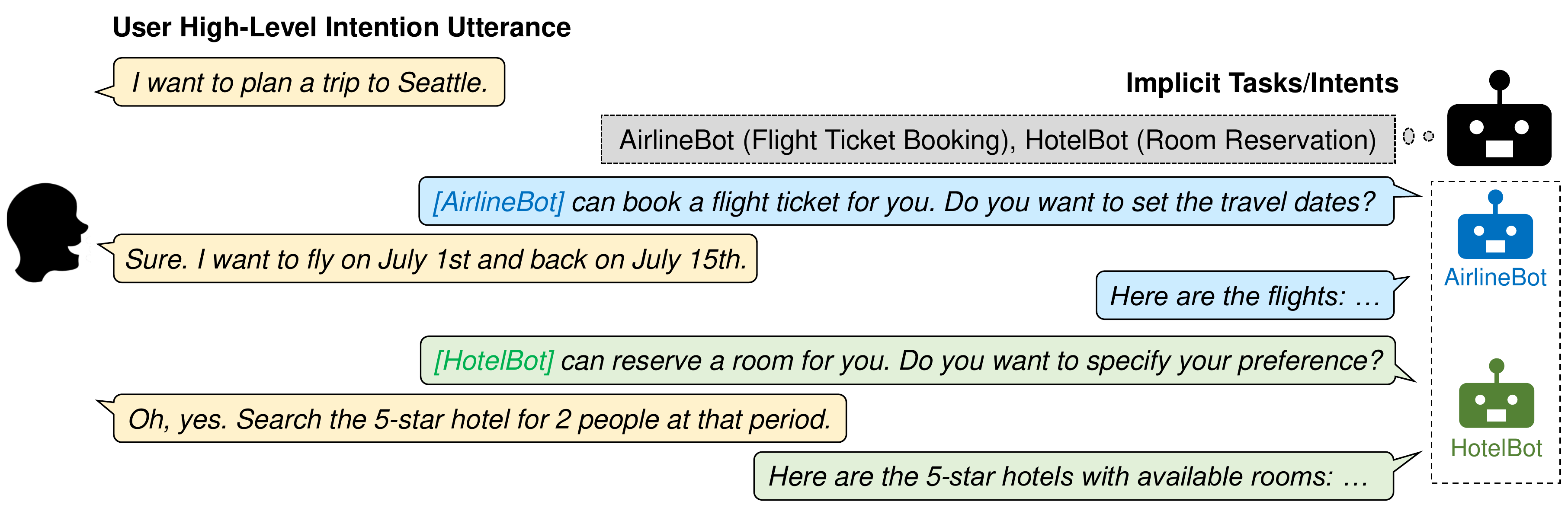}
  \caption{Illustration of a multi-task dialogue example. }
  \label{fig:dialog_ex}
\end{figure*}

Intelligent assistants have become increasingly popular in recent years, but they require users to \emph{explicitly} describe their tasks within a \emph{single} domain.
Yet, the exploration of gradually guiding users through individual task-oriented dialogues has been relatively limited~\cite{chiu2022salesbot}.
This limitation is amplified when tasks extend across multiple domains, compelling users to interact with numerous bots to accomplish their goals~\cite{sun2016appdialogue}.
For instance, planning a trip might involve interacting with one bot for flight booking and another for hotel reservation, each requiring distinct, task-specific intentions like ``{\it Book a flight ticket}'' to activate the corresponding bot, such as an airline bot.
In contrast, human assistants can manage high-level intentions spanning \emph{multiple} domains, utilizing commonsense knowledge. This approach renders conversations more pragmatic and efficient, reducing the user's need to deliberate over each task separately. To overcome this limitation of current intelligent assistants, we present a flexible framework capable of recommending task-oriented bots within a multi-domain dialogue system, leveraging commonsense-inferred \emph{implicit} intents as depicted in Figure~\ref{fig:dialog_ex}.

\paragraph{Multi-Domain Realization}
\citet{sun2016appdialogue} pinpointed the challenges associated with a multi-domain dialogue system, such as 1) comprehending single-app and multi-app language descriptions, and 2) conveying task-level functionality to users. They also gathered multi-app data to encourage research in these directions. The HELPR framework~\cite{sun2017helpr} was the pioneering attempt to grasp users' multi-app intentions and consequently suggest appropriate individual apps. Nevertheless, previous work focused on understanding individual apps based on high-level descriptions exclusively through user behaviors, necessitating a massive accumulation of personalized data. 
Due to the lack of paired data for training, our work leverages external commonsense knowledge to bridge the gap between high-level utterances and their task-specific bots. This approach enables us to consider a broad range of intents for improved generalizability and scalability.

\paragraph{Commonsense Reasoning}
Commonsense reasoning involves making assumptions about the nature and essence of typical situations humans encounter daily. These assumptions encompass judgments about the attributes of physical objects, taxonomic properties, and individuals' intentions. Existing commonsense knowledge graphs such as ConceptNet \citep{bosselut2019comet}, \textsc{Atomic} \citep{sap2019atomic}, and TransOMCS \citep{zhang2021transomcs} facilitate models to reason over human-annotated commonsense knowledge.
This paper utilizes a generative model trained on \textsc{Atomic$^{20}_{20}$} \citep{hwang2021comet} to predict potential intents linking given user high-level utterances with corresponding task-oriented bots. The inferred intents can activate the relevant task-oriented bots and also serve as justification for recommendations, thereby enhancing explainability. This work is the first attempt to integrate external commonsense relations with task-oriented dialogue systems.

\paragraph{Zero-Shot Prompting} 
Recent research has revealed that large language models \citep{radford2019language, brown2020language} have acquired an astounding ability to perform few-shot tasks by using a natural-language prompt and a handful of task demonstrations as input context~\cite{brown2020language}. 
Guiding the model with interventions via an input can render many downstream tasks remarkably easier if those tasks can be naturally framed as a cloze test problem through language models.
As a result, the technique of prompting, which transposes tasks into a language model format, is increasingly being adopted for different tasks~\cite{zhao2021calibrate,schick2021exploiting}. 
Without available data for prompt engineering~\cite{shin2020autoprompt}, we exploit the potential of prompting for bot recommendation in a zero-shot manner. 
This strategy further extends the applicability of our proposed framework and enables it to accommodate a wider variety of user intents and tasks, thus contributing to a more versatile and efficient multi-domain dialogue system.

\section{Framework}

\begin{figure*}[t!]
  \centering
  \includegraphics[width=\linewidth]{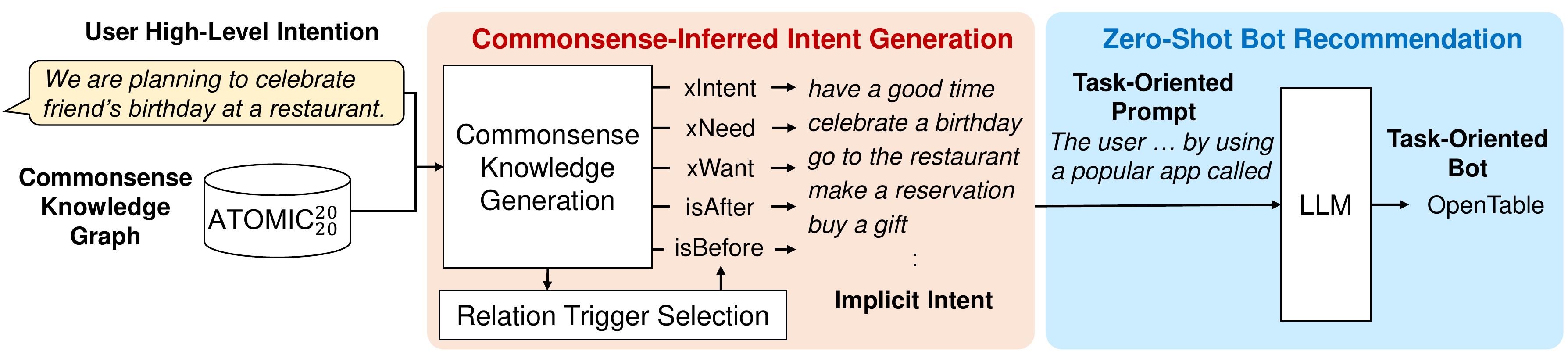}
  \caption{Our zero-shot framework for triggering task-oriented bots via the commonsense-inferred prompts.}
  \label{fig:framework_design}
\end{figure*}

Figure \ref{fig:framework_design} illustrates our proposed two-stage framework, which consists of: 1) a commonsense-inferred intent generator, and 2) a zero-shot bot recommender. 
Given a user's high-level intention utterance, the first component focuses on generating implicit task-oriented intents. The second component then utilizes these task-specific intents to recommend appropriate task-oriented bots, considering the bots' functionality through a large pre-trained language model.

\subsection{Commonsense-Inferred Intent Generation} 

The commonsense-inferred implicit intents function not only as prompts for bot recommendation but also as rationales for the suggested bots, thereby establishing a solid connection between the high-level intention and task-oriented bots throughout the conversation. For instance, the multi-domain system shown in Figure~\ref{fig:dialog_ex} recommends not only the \emph{AirlineBot} but also describes its functionality---``{\it can book a flight ticket}''---to better convince the user about the recommendation.

\subsubsection{Relation Trigger Selection}\label{subsec:relation-generation}

\textsc{Atomic$^{20}_{20}$} is a commonsense knowledge graph featuring commonsense relations across three categories: social-interaction, event-centered, and physical-entity relations, all of which concern situations surrounding a specified event of interest. 
Following \citet{hwang2021comet}, we employ a BART model~\cite{lewis2020bart} pre-trained on \textsc{Atomic$^{20}_{20}$} to generate related entities and events based on the input sentence.
However, despite having a total of 23 commonsense relations, not all are suitable for inferring implicit intents in assistant scenarios. 
We utilize AppDialogue data \cite{sun2016appdialogue} to determine which commonsense relations can better trigger the task-specific intents. 
Given a high-level intention description $u_i$ and its task-specific sentences $s_{ij}$, we calculate the trigger score of each relation $r$ as an indicator of its suitability as a trigger relation.
\begin{equation}
T(r) = \sum_i \sum_ j P_{BART}([u_i, r, s_{ij}]),
\end{equation}
where $P_{BART}([u_i, r, s_{ij}])$ represents the probability of the sentence beginning with the high-level user description $u_i$, followed by a relation trigger $r$, and the corresponding task-specific sentences $s_{ij}$. 
By summing up multiple task-specific sentences over $j$ and all samples over $i$, a higher $T(r)$ implies that the relation $r$ can better trigger implicit task-oriented intents in assistant scenarios.

We identify a total of five relations with the highest $T(r)$ and present their definitions~\cite{sap2019atomic} in Table~\ref{tab:atomic-relations}.
These relations are also reasonable from a human perspective to trigger implicit user intents.

\begin{table}[t!]
\centering\small
\begin{tabular}{ clp{5.0cm}  }
\toprule
& \bf Relation & \bf Definition\\
\toprule
\multirow{7}{*}{\rotatebox[origin=c]{90}{Social}} & \bf xIntent & the likely intent or desire of an agent (X) behind the execution of an event\\
& \multicolumn{2}{p{6.5cm}}{``X gives Y gifts'' $\rightarrow$ X wanted ``to be thoughtful''} \\
\cmidrule{2-3}
& \bf xNeed & a precondition for X achieving the event \\
& \multicolumn{2}{p{6.5cm}}{``X gives Y gifts'' $\rightarrow$ X must first ``buy the presents''} \\
\cmidrule{2-3}
& \bf xWant & post-condition desires on the part of X\\
& \multicolumn{2}{p{6.5cm}}{``X gives Y gifts'' $\rightarrow$ X may also desire ``to hug [Y]''} \\
\midrule
\multirow{4}{*}{\rotatebox[origin=c]{90}{Event}} & \bf isAfter & events that can precede an event\\
& \multicolumn{2}{p{6.5cm}}{``X is in a hurry to get to work'' $\rightarrow$ ``X wakes up late''}\\
\cmidrule{2-3}
& \bf isBefore & events that can follow an event\\
& \multicolumn{2}{p{6.5cm}}{``X is in a hurry to get to work'' $\rightarrow$ ``X drives too fast''}\\
\bottomrule
\end{tabular}
\caption{Selected relations from ATOMIC$^{20}_{20}$.}
\label{tab:atomic-relations}
\end{table}

\subsubsection{Commonsense Knowledge Generation}
\label{subsec:commonsense-generation}

Given the selected relations $R = \{r_1, r_2, ..., r_5\}$, where $r_i$ represents the $i$-th relation from \textbf{\{xIntent, xNeed, xWant, isAfter, isBefore\}}, we concatenate each relation with a user utterance $u$ to serve as the context input for our pre-trained BART model: 
\[\verb|<s>|\;\;\; u \ \ r_i \ \ \verb|[GEN]| \;\;\; \verb|</s>|,\]
where \verb|<s>| and \verb|</s>| are special tokens in BART, and \verb|[GEN]| is a unique token employed during the pre-training of BART to initiate the commonsense-related events. 
BART accepts this input and decodes the commonsense events into implicit task-oriented intents $Y = {y^1_{1:k}, y^2_{1:k}, ..., y^5_{1:k}}$, where $y^i_k$ denotes the $k$-th generated commonsense event of the relation $r_i$.

\subsection{Zero-Shot Bot Recommendation}
With the inferred intents, the second component aims to recommend appropriate bots capable of executing the anticipated tasks. To pinpoint the task-specific bots based on the required functionality, we leverage the remarkable capacity of a large pre-trained language model, assuming that app descriptions form a part of the pre-trained data.

\subsubsection{Pre-trained Language Model}\label{subsec:prompting}

The language model used in this study is GPT-J 6B\footnote{\url{https://huggingface.co/EleutherAI/gpt-j-6B}}, an GPT-3-like causal language model trained on the Pile dataset\footnote{\url{https://pile.eleuther.ai/}}~\cite{radford2019language}, a diverse, open-source language modeling dataset that comprises 22 smaller, high-quality datasets combined together. Making the assumption that app descriptions in mobile app stores are incorporated in the pre-training data, we exploit the learned language capability to suggest task-oriented bots based on the given intents.

\begin{table*}[t!]
  \centering\small
  \begin{tabular}{lcccc}
    \toprule
    \bf Method   & \bf Precision & \bf Recall & \bf F1 & \bf Human Score (Mean$\pm$STD) \\
    \midrule
    1-Stage Prompting Baseline & 30.3 & 20.6 & 23.7 & 1.73$\pm$1.03\\
    2-Stage Prompting (GPT-3) & 28.6 & \bf 41.7 & 31.8 & 2.11$\pm$0.46\\
    Proposed 2-Stage (COMeT) & \bf 36.0 & 35.7 & \bf 32.9 & \bf 2.18$\pm$0.34\\
    Proposed 2-Stage (COMeT) w/o Reasons & - & - & - &  2.15$\pm$0.35 \\
    \midrule
    \it Gold & & & & \it 2.44$\pm$0.27\\
    \bottomrule
  \end{tabular}
  \caption{Evaluation scores (\%).}
 \label{tab:score-table1}
\end{table*}

\begin{table*}[t!]
  \centering
  \small
  \begin{tabular}{p{2.5cm}p{11cm}c}
    \toprule
    \bf User Input & \it We are planning to celebrate friend's birthday at a restaurant in [City]. & \bf Score\\
    {\bf User-labeled} & Line (Communication), Google Maps (Maps \& Navigation), Calendar (Productivity) & 2.25\\ 
    {\bf 1-Stage Prompting} & Tinder (Lifestyle), Grindr (Lifestyle) & 1.83\\ 
    \bf 2-Stage Prompting & Zomato can help  to book the restaurant in advance. & 2.00\\
    & WhatsApp can find out about their contact information. & \\
    \midrule
    \bf Proposed 2-Stage & WhatsApp can help have a good time and to celebrate a friend's birthday & \bf 2.67\\
    & OpenTable can help book a table at the restaurant and go to the restaurant.\\
    \midrule
    {\bf  w/o Reasons} & WhatsApp (Communication), OpenTable (Food \& Drink) & 2.17 \\ 
    \bottomrule
    \end{tabular}
  \caption{Generated results for given user high-level descriptions.}
 \label{tab:qual}
\end{table*}

\subsubsection{Prompting for Bot Recommendation}\label{subsec:app-generation}

To leverage the pre-trained language capability of GPT-J, we manually design prompts for each relation type. 
For social-interaction relations, the prompt is formulated as ``\textit{The user $r_i$ $y^i_{1:k}$ by using a popular app called}''.
For instance, Figure~\ref{fig:framework_design} generates a prompt ``{\it The user needs to go to the restaurant and make the reservation by using a popular app called}''.
For event-centered relations, we simply concatenate the generated events and app-prompt to trigger the recommended task-oriented apps/bots.

\section{Experiments}

To evaluate the zero-shot performance of our proposed framework, we collected a test set specific to our multi-domain scenarios. We recruited six volunteers who were knowledgeable about the target scenarios to gather their high-level intention utterances along with the associated task-oriented bots. 
Upon filtering out inadequate data, our test set incorporated a total of 220 task-oriented bots and 92 high-level utterances, each linked with an average of 2.4 bots. The number of bot candidates considered in our experiments is 6,264, highlighting the higher complexity of our tasks.

Our primary aim is to connect a high-level intention with its corresponding task-oriented bot recommendation by leveraging external commonsense knowledge. 
Therefore, we assess the effectiveness of the proposed methodology and compare it with a 1-stage prompting baseline using GPT-J to maintain fairness in comparison. 
For this baseline, we perform simple prompting on the user's high-level utterance concatenating with a uniform app-based prompt: ``{\it so I can use some popular apps called}.''
In response to these context prompts, GPT-J generates the associated (multiple) app names, serving as our baseline results.

To further investigate whether our proposed commonsense-inferred implicit intent generator is suitable for our recommendation scenarios, we introduce another 2-stage prompting baseline for comparison. 
Taking into account that contemporary large language models exhibit astonishing proficiency in commonsense reasoning, we substitute our first component with the state-of-the-art GPT-3~\cite{brown2020language} to infer implicit intents, serving as another comparative baseline.

\subsection{Automatic Evaluation Results} \label{subsec:experiment-result}


Considering that multiple bots can fulfill the same task (functionality), we represent each app by its category as defined on Google Play, then compute precision, recall, and F1 score at the \emph{category} level. 
This evaluation better aligns with our task objective; for instance, both ``\emph{WhatsApp}'' and ``\emph{Line}'' belong to the same category---``communication'' as demonstrated in Table~\ref{tab:qual}.



Table~\ref{tab:score-table1} presents that the 2-stage methods significantly outperform the 1-stage baseline, suggesting that commonsense knowledge is useful to bridge high-level user utterances with task-oriented bots. 
Further, our proposed approach, which leverages external commonsense knowledge, achieves superior precision over GPT-3, a quality that is more important in recommendation scenarios.
The reason is that GPT-3 may generate hallucinations for inferring more diverse but may not suitable intents.

\subsection{Human Evaluation Results}
\label{subsec:human-evaluation}

Given that our goal can be interpreted as a recommendation task, the suggested bots different from user labels can be still reasonable and useful to users.
Therefore, we recruited crowd workers from Amazon Mechanical Turk (AMT) to evaluate the relevance of each recommended result given its high-level user utterance.
Each predicted bot or app is assessed by three workers on a three-point scale: \textbf{irrelevant} (1), \textbf{acceptable} (2), and \textbf{useful} (3).
The human-judged scores are reported in the right part of Table~\ref{tab:score-table1}, and our proposed framework achieves the average score of 2.18, implying that most recommended tasks are above acceptable.
Compared with the 1-stage baseline with a score below 2, it demonstrates that commonsense inferred implicit intents can more effectively connect the reasonable task-oriented bots.
Considering that the score of 2-stage prompting is also good, we report the pair-wise comparison in Table~\ref{tab:human}, where we can see that humans prefer ours to 2-stage prompting baseline for 57\% of the data.

\begin{table}[t!]
  \centering\small
  \begin{tabular}{lccr}
    \toprule
    \bf Method  & \bf Win & \bf Lose & \bf Tie\\
    \midrule
    Ours vs. 2-Stage Prompt (GPT-3)  & 57.6 & 40.2 & 2.2\\
    Ours vs. Ours w/o Reasons & 55.1 & 38.8 & 6.1\\ 
    \bottomrule
  \end{tabular}
  \caption{Pair-wise human preference results (\%). }
 \label{tab:human}
\end{table}

In additon to simply suggesting task-oriented bots, providing the rationale behind their recommendation could help users better judge their utility. 
Within our proposed framework, the commonsense-inferred implicit intents, which are automatically generated by the first component, can act as the explanations for the recommended task-oriented bots, as illustrated in Table~\ref{tab:qual}. 
Consequently, we provide these rationales alongside the recommended results using the predicted intents and undergo the same human evaluation process.
Table~\ref{tab:human} validates that providing these justifications results in improved performance from a human perspective, further suggesting that commonsense-inferred intents are useful not only for prompting task-oriented bots but also for generating human-interpretable recommendation.

\begin{table}[t!]
  \centering
  \small
  \begin{tabular}{p{0.9cm}p{5.7cm}c}
    \toprule
    \multicolumn{2}{c}{\bf Generated Intent Example} \\
    \midrule
    \bf Input & \it My best friend likes pop music. \\
    \midrule
    COMet & Want $\rightarrow$ to listen to music\\
    & Intent $\rightarrow$ to be entertained\\
    & Need $\rightarrow$ to listen to music\\
    GPT-3 & Want $\rightarrow$ to get her tickets to see Justin Bieber for her birthday \\
    & Intent $\rightarrow$ to buy her a CD by Taylor Swift for her birthday \\
    & Need $\rightarrow$ to find songs that are pop and appropriate for her \\
    \midrule
    \bf Input & \it I am looking for a job. \\
    \midrule
    COMet & Want $\rightarrow$ to apply for a job\\
    & Intent $\rightarrow$ to make money\\
    & Need $\rightarrow$ to apply for a job\\
    GPT-3 & Want $\rightarrow$ to learn more \\
    & Intent $\rightarrow$ to apply for a job \\
    & Need $\rightarrow$ to update my resume \\
    \bottomrule
    \end{tabular}
  \caption{Generated commonsense-inferred intents.}
 \label{tab:intent-output}
\end{table}

\section{Discussion}

Table~\ref{tab:intent-output} showcases the implicit intents generated by our proposed COMeT generator and GPT-3. It is noteworthy that GPT-3 occasionally produces hallucinations, which can render the recommended bots unsuitable.
For instance, given the text prompt ``{\it My best friend likes pop music.}'', GPT-3 infers an intent to ``{\it buy a ticket to see Justin Bieber}'', which may not align accurately with the user's need.

Hence, our experiments reveal that while the 2-stage prompting achieves higher recall, its precision is lower. As our objective is to recommend reasonable task-specific bots, a higher precision is more advantageous in our scenarios.





\section{Conclusion}

This paper introduces a pioneering task centered around recommending task-oriented dialogue systems solely based on high-level user intention utterances. 
The proposed framework leverages the power of commonsense knowledge to facilitate zero-shot bot recommendation.
Experimental results corroborate the reasonability of the recommended bots through both automatic and human evaluations. 
Experiments show that the recommended bots are reasonable for both automatic and human evaluation, and the inferred intents can provide informative and interpretable rationales to better convince users of the recommendation for  practical usage.
This innovative approach bridges the gap between user high-level intention and actionable bot recommendations, paving the way for a more intuitive and user-centric conversational AI landscape.

\section*{Limitations}

This paper acknowledges three main limitations: 1) the constraints of a zero-shot setting, 2) an uncertain generalization capacity due to limited data in the target task, and 3) the longer inference time required by a large language model.

Given the absence of data for our task and the complexity of the target scenarios, collecting a large dataset for supervised or semi-supervised learning presents a significant challenge.
As the first approach tackling this task, our framework performs the task in a zero-shot manner, but is applicable to fine-tuning if a substantial dataset becomes available. 
Consequently, we expect that future research could further train the proposed framework using supervised learning or fine-tuning, thereby enhancing the alignment of inferred implicit intents and recommended bots with training data.
This would expand our method to various learning settings and validate its generalization capacity.

Conversely, the GPT-J model used for recommending task-oriented bots is considerably large given academic resources, thereby slowing down inference speed.
To mitigate this, our future work intends to develop a lightweight student model that accelerates the prompt inference process. Such a smaller language model could not only expedite the inference process to recommend task-oriented bots but also be conveniently fine-tuned using collected data.

Despite these limitations, this work can be considered as the pioneering attempt to leverage commonsense knowledge to link task-oriented intents. The significant potential of this research direction is evidenced within this paper.

\section*{Ethics Statement}

This work primarily targets the recommendation of task-oriented bots, necessitating a degree of personalization. To enhance recommendation effectiveness, personalized behavior data may be collected for further refinement. 
Balancing the dynamics between personalized recommendation and privacy is a critical consideration. 
The data collected may contain subjective annotations, and the present paper does not dive into these issues in depth. 
Future work should address these ethical considerations, ensuring an balance between personalized recommendations and privacy preservation.

\section*{Acknowledgements}
We thank the reviewers for their insightful comments. This work was financially supported by the Young Scholar Fellowship Program by the National Science and Technology Council (NSTC) in Taiwan, under Grants 111-2222-E-002-013-MY3 and 111-2628-E-002-016.

\bibliography{anthology,arxiv}
\bibliographystyle{acl_natbib}


\appendix

\section{Implementation Details}
In our zero-shot bot recommendation experiments, which are evaluated using Android apps based on RICO data~\cite{deka2017rico}, we append the phrase ``{\it in Android phone}'' to all prompts. 
This helps guide the resulting recommendations. 
Task-oriented prompts are fed into GPT-J to generate token recommendations for bots/apps, such as ``{\it OpenTable}'', an Android app, which aligns better with our evaluation criteria.

In the 2-stage prompting baseline, our prompts for GPT-3, designed to generate commonsense-related intents, are coupled with our selected relations to ensure a fair comparison. These prompts are outlined in Table~\ref{tab:gpt3-prompt}.

\begin{table}[t!]
    \centering\small
    \begin{tabular}{clp{4.0cm}}
    \toprule
    & \bf Relation & \bf GPT-3 Prompt\\
    \toprule
    \multirow{3}{*}{\rotatebox[origin=c]{90}{Social}} & \bf xIntent & so I intend\\
    \cmidrule{2-3}
    & \bf xNeed & so I need \\
    \cmidrule{2-3}
    & \bf xWant & so I want\\
    
    \midrule
    \multirow{2.5}{*}{\rotatebox[origin=c]{90}{Event}} & \bf isAfter & Before, the user needs to \\
    \cmidrule{2-3}
    & \bf isBefore & After, the user needs to\\
    \bottomrule
    \end{tabular}
    \vspace{2mm}
    \caption{Designed prompts of GPT-3. The prompts are converted from selected relations of ATOMIC$^{20}_{20}$ for a fair comparison.}
    \label{tab:gpt3-prompt}
\end{table}

\begin{figure*}[t!]
  \centering
  \includegraphics[width=1.\linewidth]{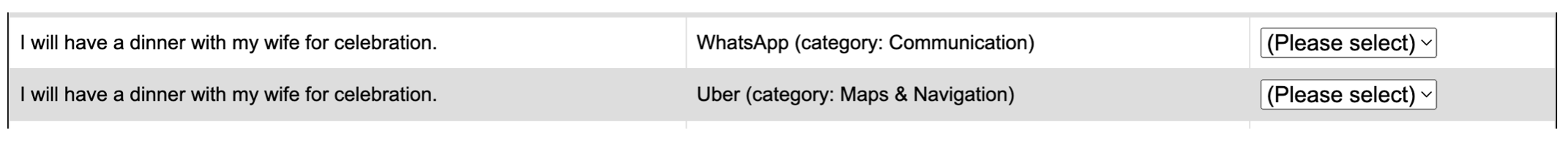}
  \vspace{-6mm}
  \caption{An annotation screenshot of annotating the recommended apps/bots on the Amazon Mechanical Turk, where the results may come from the ground truth, the baseline, or the proposed method.}
  \label{fig:annotation1}
  \vspace{-2mm}
\end{figure*}

\begin{figure*}[t!]
  \centering
  \includegraphics[width=1.\linewidth]{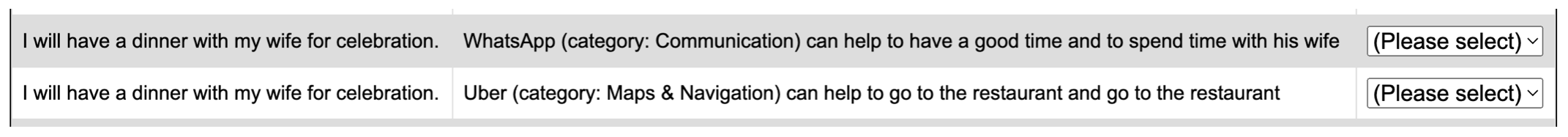}
  \vspace{-6mm}
  \caption{An annotation screenshot of annotating the recommended apps/bots together with the predicted intents as reasons on the Amazon Mechanical Turk.}
  \label{fig:annotation2}
  \vspace{-2mm}
\end{figure*}

\section{Reproducibility}

To enhance reproducibility, we release our data and code.
Detailed parameter settings employed in our experiments are as follows.

In commonsense knowledge generation, we apply beam search during generation, setting \emph{beam\_size=10}.
In prompting for bot recommendation, a sampling strategy is implemented during recommendation generation, with \emph{max\_length=50}, \emph{temperature=0.01}, and \emph{top\_p=0.9}.

\section{Crowdsourcing Interface}
\label{sec:amt}
Figure~\ref{fig:annotation1} and \ref{fig:annotation2} display annotation screenshots for both types of outputs. Workers are presented with a recommendation result from 1) user-labeled ground truth, 2) the baseline, and 3) our proposed method. Note that results accompanied by reasons originate only from our proposed method.

\section{Qualitative Analysis}
\label{sec:qual}

Table~\ref{tab:qualitative} features additional examples from our test set, highlighting our method's ability to use commonsense knowledge to recommend more appropriate apps than the baseline, and broaden user choices.

In the first example, our method discerns the user's financial needs and suggests relevant financial apps such as \emph{Paypal}. Conversely, the baseline method could only associate the user's needs with communication apps like \emph{WeChat}, possibly influenced by the term \emph{friend} in the high-level description.

In the second example, our method infers potential user intents about checking their bank account and purchasing a new notebook, thus recommending \emph{Paypal} for bank account management and \emph{Amazon} for shopping.

In the third example, the user mentions having a tight schedule. Hence, our method suggests \emph{Uber} to expedite the user's commute to the movie theater or \emph{Netflix} for instant access to movies.

\begin{table*}[t!]
  \centering
  \small
  \begin{tabular}{p{1.7cm}p{13.1cm}}
    \toprule
    \multicolumn{2}{c}{\bf Data Example} \\
    \midrule
    \bf User Input & \it Check if my friend sent the money to me. \\
    {\bf User-labeled} & Bank (Finance), Messenger (Communication)\\ 
    {\bf Baseline} & WhatsApp (Communication), WeChat (Communication)\\ 
    {\bf Proposed} & Google Wallet (Finance), WhatsApp (Communication), Paypal (Finance)\\ 
    {\bf Reasons} &
    Google Wallet can help check if the money was sent to the right place and check if the money was sent to the correct place\\
    & WhatsApp can help find out where the money came from and find out who sent the money\\
    & Paypal can help to give the money to my friend and to give the money to the person who sent it to me\\
    \midrule
    \bf User Input & \it My notebook was broken. I need to get a new one. Check how much money is left in my account. \\
    {\bf User-labeled} & Shopee (Shopping)\\ 
    {\bf Baseline} & Google Play (Google Play)\\ 
    {\bf Proposed} & Google Play (Google Play), Amazon (Shopping), Mint (Tools), Paypal (Finance)\\ 
    \bf Reasons & Google Play can help to buy a new one and to buy a new notebook.\\
    & Amazon can help to buy a new one and find out how much money is left.\\
    & Mint can help to buy a new one and to buy a new notebook.\\
    & PayPal can help my credit card is maxed out and my credit card is maxed out and I can't afford a new one.\\
    \midrule
    \bf User Input & \it I really like watching movie, but my schedule is so tight. \\
    {\bf User-labeled} & Calendar (Productivity), Movies (Entertainment)\\ 
    {\bf Baseline} & MovieBox (Entertainment)\\ 
    {\bf Proposed} & WhatsApp (Communication), Netflix (Entertainment), Youtube (Media), Uber (Maps \& Navigation) \\ 
    \bf Reasons & WhatsApp can help to be entertained and to have fun. \\
    & Netflix can help find a movie to watch and find a movie to watch.\\
    & Youtube can help go to the movies and to find a movie to watch.\\
    & Uber can help when you have a lot of work to do and have to go to work.\\
    \bottomrule
    \end{tabular}
  \caption{Generated results for given user high-level descriptions.}
 \label{tab:qualitative}

\end{table*}

\end{document}